\documentclass[letterpaper, 10 pt, conference]{ieeeconf}  

\IEEEoverridecommandlockouts                              

\usepackage{amsmath} 
\usepackage{graphicx}
\usepackage{threeparttable}

\title{\LARGE \bf
Enhancing Neural Radiance Fields with Depth and Normal Completion Priors from Sparse Views
}

\author{Jiawei Guo$^{1}$, HungChyun Chou$^{1}$ and Ning Ding$^{1}$
\thanks{*This work was supported by National Key R\&D Program of China No. U2013202 and the Guangdong Basic and Applied Basic Research Foundation under Grant No. 2022A1515011139.}
\thanks{$^{1}$Authors 1 are with Special Robot Center, Shenzhen Institute of Artificial Intelligence and Robotics for Society, 14-15F, Tower G2, Xinghe World, Rd Yabao, Longgang District, Shenzhen, Guangdong, 518129, China.
        {\tt\small jiaweiguo@cuhk.edu.cn}, {\tt\small jeffrey.guojiawei@gmail.com} and
        {\tt\small zhouhongjun@cuhk.edu.cn}}%
}

\begin{document}

\maketitle
\thispagestyle{empty}
\pagestyle{empty}

\begin{abstract}

 Neural Radiance Fields (NeRF) are an advanced technology that creates highly realistic images by learning about scenes through a neural network model. However, NeRF often encounters issues when there are not enough images to work with, leading to problems in accurately rendering views. The main issue is that NeRF lacks sufficient structural details to guide the rendering process accurately. To address this, we proposed a Depth and Normal Dense Completion Priors for NeRF (CP\_NeRF) framework. This framework enhances view rendering by adding depth and normal dense completion priors to the NeRF optimization process. Before optimizing NeRF, we obtain sparse depth maps using the Structure from Motion (SfM) technique used to get camera poses. Based on the sparse depth maps and a normal estimator, we generate sparse normal maps for training a normal completion prior with precise standard deviations. During optimization, we apply depth and normal completion priors to transform sparse data into dense depth and normal maps with their standard deviations. We use these dense maps to guide ray sampling, assist distance sampling and construct a normal loss function for better training accuracy. To improve the rendering of NeRF's normal outputs, we incorporate an optical centre position embedder that helps synthesize more accurate normals through volume rendering. Additionally, we employ a normal patch matching technique to choose accurate rendered normal maps, ensuring more precise supervision for the model. Our method is superior to leading techniques in rendering detailed indoor scenes, even with limited input views.

\end{abstract}

\section{INTRODUCTION}

The task of view synthesis is crucial for Human-Computer Interactions in virtual settings, facilitating virtual navigation across diverse buildings, indoor scenes, or simulations. Among various view synthesis methods, Neural Radiance Fields (NeRF) has emerged as a particularly influential technique. NeRF constructs a scene using an implicit volumetric representation, enabling rendering from novel viewpoints. It begins by sampling coordinates along camera rays from different angles. NeRF then employs a multilayer perceptron network (MLP) to learn a differentiable 5D implicit function. This function produces a single volume density and view-dependent RGB color from a singular 5D coordinate input. Finally, NeRF applies volume rendering to integrate all colors and densities into a cohesive 2D image.

While NeRF excels in creating photo-realistic synthesized views, it demands high-quality inputs. The images of the scene must be captured when it is nearly static, and there should be enough images to ensure significant overlap between them. However, having too many images can prolong training time of the NeRF, sometimes taking days to model a single scene on one GPU. This slow training pace is mainly due to the intensive ray-casting operations, which become more demanding as input resolution increases. Typically, NeRF relies solely on color loss for training convergence. This leads to poor view synthesis and geometry estimation in scenes with large low-texture areas or where input views have inconsistent color values.

To enhance view synthesis quality, we introduce a framework that incorporates depth and normal completion priors. Initially, we extract sparse depth information using the Structure from Motion (SfM) technique used to obtain camera positions. Based on the sparse depth maps and a normal estimator, sparse normal maps are derived for training a normal completion prior with precise standard deviations. Subsequently, we apply depth and normal completion priors to convert the sparse data into dense depth and normal maps. These dense maps then serve to refine NeRF training by optimizing ray sampling, guiding distance sampling and constructing a normal loss function, leading to increased precision. Furthermore, to improve the rendering of normal maps, we incorporate a pose embedder with volume rendering techniques. We also utilize a normal patch matching method to evaluate the quality of the rendered normal maps, providing the model with more accurate supervision signals. Our methodology is validated through experiments on the ScanNet datasets [1], demonstrating its capability to significantly improve both view synthesis and geometry estimation.

\section{RELATED WORKS}

\subsection{Novel-View Synthesis}

Novel-view synthesis is a technique that creates new perspectives of a scene from a limited number of existing images, crucial for immersive experiences in virtual environments. Traditionally, when there is an abundance of scene images, light field interpolation [2] can be used to craft new views. However, in cases where image samples are sparse, incorporating geometric data becomes essential for synthesizing these novel views. Research has been conducted on integrating geometric information [3, 4] within various frameworks to achieve this. Additionally, some approaches employ 3D shape representations like voxel grids [5] or multiplane images [6] to generate new viewpoints. NeRF, a recent innovation, uses neural radiance fields to represent scenes implicitly. NeRF is defined by one MLP and rendered through volume rendering techniques. NeRF has shown remarkable success in creating new views for complex scenes, establishing itself as a leading method in scene representation for novel-view synthesis.

\subsection{NeRF from Few Views}

The reliance on densely-sampled images for NeRF is a significant limitation. To address this, recent innovations aim to reduce the required number of inputs. PixelNeRF [7] and metaNeRF [8] utilize data-driven priors from training scenes to fill in gaps in test scenes. But these approaches depend on having ample training data and a clear understanding of the differences between training and test scenes. MVS-NeRF introduces another strategy by integrating 3D scene information onto a plane sweep volume. This work employs 3D CNNs to recreate a neural encoding volume for precise rendering [12]. However, MVS-NeRF still demands a considerable number of images for successful training, relying solely on RGB value supervision for model convergence. Alternatively, leveraging priors from related tasks like depth estimation [9], semantic prediction [10], or surface normal estimation [11] offers a potential route to lessen the input requirement. These priors provide auxiliary information to guide the rendering process of the NeRF.

\subsection{NeRF with Depth and Normal}

NerfingMVS [9] leverages a depth network to transform sparse depth data from Multi-View Stereo (MVS) into dense depth data, enhancing NeRF training by improving ray sampling. DietNeRF [10] introduces a novel approach by incorporating a semantic consistency loss, applied in a high-level feature space. This method can ensure the rendering accuracy from any viewpoint, not just predetermined ones. Depth-supervised NeRF [13] uses sparse depth information from SfM to create dense depth maps for training supervision. To maintain depth accuracy, the reprojection error weights are applied. Another method [14] utilizes dense depth priors to construct a depth loss during NeRF optimization, with uncertainties from a depth completion network.

Building on these concepts, we propose a framework that uses both dense depth and normal priors. The depth prior is trained from sparse maps during camera pose estimation via SfM. And the normal prior is trained from normal outputs of a normal estimation model. Inspired by SS-NeRF [11], we integrate a pose embedder to synthesize more accurate normal maps through volume rendering. Additionally, we employ a normal patch matching technique to choose accurate rendered normal maps, ensuring more precise supervision for the model. Our method outperforms the baseline [14] in rendering high-quality views in complex indoor scenes with few input images. This demonstrates its effectiveness in enhancing view synthesis capabilities of NeRF.

\section{PROPOSED METHOD}

\begin{figure*}
\centering
\includegraphics[width=\linewidth]{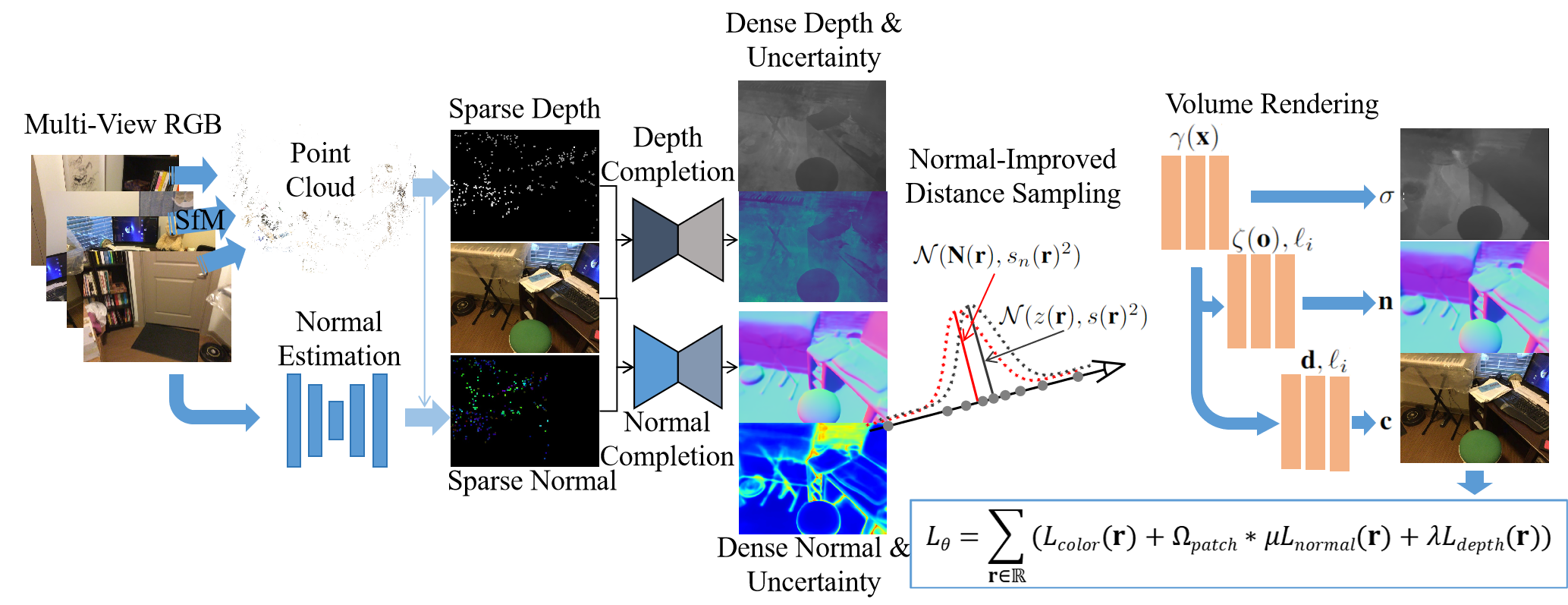}
\caption{The proposed framework.}\label{fig:Framework}
\end{figure*}

Our proposed method achieves photo-realistic indoor novel view synthesis based on a sparse set of image inputs $\{I_i\}$, $I_i \in [0, 1]^{H \times W \times 3}$, $i \in [0, N-1]$. We first exploit SfM to generate camera poses $\textbf{p}_i \in \mathbf{R}^6$, intrinsic values $K_i \in \mathbf{R}^{3 \times 3}$, and sparse depth maps $Z_{\rm sparse\_depth} \in [0, d_  f]^{H \times W}$ for every image input. $d_f$ signifies the farthest point in volume rendering. Concurrently, sparse normal maps $Z_{\rm sparse\_normal} \in [0, 1]^{H \times W \times 3}$ are derived from a normal estimation model based on the sparse depth maps.

\begin{equation}\label{eq:DepthCompletion}
\left\{
\begin{aligned}
& [Z_{\rm dense\_depth}, S_{\rm depth}] = D_\theta(I_i, Z_{\rm sparse\_depth})
\\
& [Z_{\rm dense\_normal}, S_{\rm normal}] = D_\phi(I_i, Z_{\rm sparse\_normal})
\end{aligned}
\right.
\end{equation}

Drawing inspiration from prior work [14], we apply a depth completion technique to transform sparse depth maps into dense depth maps. Similarly, we enrich sparse normal maps to dense ones with accurate standard deviations using a normal completion approach. These dense maps are then integrated into NeRF’s training process. The dense normal maps assist the dense depth maps to facilitate improved ray distance sampling for more efficient NeRF optimization, where a normal loss is introduced to refine the process further. Meanwhile, we incorporate an optical centre position embedder within the volume rendering phase to aid in generating more precise normal outputs. Additionally, to ensure the effectiveness of the rendered normal maps for the normal loss, we employ a normal patch matching technique. This method effectively filters out rendered normals that contain excessive noise, thereby enhancing the quality of the normal supervision signals used in the optimization process.

\subsection{Depth and Normal Completion with Uncertainty} 

\textbf{Model Architecture:} The process of our depth and normal dense completion technique is illustrated in Fig. 1. Initially, we acquire sparse depth maps from a limited collection of RGB images through SfM reconstruction used to obtain camera poses. Concurrently, sparse normal maps are produced for training a normal completion prior with precise standard deviations, using a normal estimation model assisted by the sparse depth maps. Next, we establish two separate networks: a depth prior network $D_\theta$ and a normal prior network $D_\phi$, which trained from the obtained sparse depth and normal maps, respectively. The prior networks $D_\theta$ and $D_\phi$ are structured as convolutional networks using a ResNet-based encoder-decoder with a Convolutional Spatial Propagation Network (CSPN) module and skip connection layers. These networks are then responsible for generating dense depth maps $Z_{\rm dense\_depth}$ and dense normal maps $Z_{\rm dense\_normal}$, complete with their standard deviations $S_{\rm depth}$ and $S_{\rm normal}$ as (1).

\textbf{Model Training:} In the SfM reconstruction process, we determine the 2D positions of sparse depth points by employing a SIFT (Scale-Invariant Feature Transform) feature extractor. The 2D positions of sparse normal points are calculated by the sparse depth points. For training the depth prior network, we use sparse depth maps in grayscale format as input. Conversely, during the training of the normal prior network, we input sparse normal maps in RGB format. Throughout this training process, we assume that the outputs follow a normal distribution. To construct the training loss of the depth completion prior, we minimize the negative log likelihood of a Gaussian distribution (GNLL), effectively aligning the network's predictions with the assumed normal distribution of the output data.

\begin{equation}\label{eq:DepthCompletionLoss}
L_{\theta\_\rm prior\_GNLL} = \frac{1}{n}\sum_{j=1}^n (\mathrm{log}(s_j^2) + \frac{(z_j - z_{\mathrm{sensor}, j})^2}{s_j^2})
\end{equation}

Here, $z_j$ and $s_j$ represent the predicted depth and the standard deviation for pixel $j$, $z_{\mathrm{sensor},j}$ refers to the actual depth from sensor at $j$, and $n$ signifies the total number of valid pixels within the dense sensor depth map.

In the training process for normal completion prior, we combine an angular von Mises-Fisher distribution loss (AngMF) with the GNLL to supervise the normal completion network, inspired by SNU [16]. This integration aims to enhance the accuracy and reliability of the predicted normals by considering both their directional alignment and the probabilistic distribution of errors.
    
\begin{equation}\label{eq:NormalCompletionLoss}
\left\{
\begin{aligned}
& L_{\phi\_\rm prior\_GNLL} = \frac{1}{m}\sum_{j=1}^m (\mathrm{log}(s_{j\_\rm n}^2) + \frac{(\textbf{n}_j - \textbf{n}_{\mathrm{gt}\_j})^2}{s_{j\_\rm n}^2}) 
\\
& L_{\phi\_\rm prior\_AngMF} = \kappa_j\mathrm{cos}^{-1}(\textbf{n}_j^T\textbf{n}_{\mathrm{gt}\_j}) - \mathrm{log}(\kappa_j^2 + 1) \\ 
& \quad\quad\quad\quad\quad\quad\quad + \mathrm{log}(1 + \mathrm{exp}(-\kappa_j\pi))
\end{aligned}
\right.
\end{equation}

In this context, $\textbf{n}_j$ and $s_{j\_\rm n}$ denote the predicted normal vector and its associated standard deviation for pixel $j$, respectively. The term $\textbf{n}_{\mathrm{gt}\_j}$ refers to the ground truth normal vector at pixel $j$. $\kappa_j$ is the concentration parameter derived from $s_{j\_\rm n}$. Both $\textbf{n}_j$ and $\textbf{n}_{\mathrm{gt}\_j}$ are unit vectors and $\kappa_j$ is greater than 0. High $\kappa_j$ means that the standard deviation of the predicted normal at pixel $j$ is low. $m$ signifies the total number of valid pixels within the dense actual normal map.

\subsection{NeRF with Dense Depth and Normal Priors}

\textbf{Scene Representation:} Inspired by [14], we construct the radiance field of the scene into three separate MLPs: $F_{\theta\_1}$, $F_{\theta\_2}$, and $F_{\theta\_3}$. These MLPs are specialized to predict different aspects of the scene. $F_{\theta\_1}$ focuses on estimating the volumn density $\sigma$, $F_{\theta\_2}$ is tasked with predicting the color $\textbf{c} = [r, g, b]$, and $F_{\theta\_3}$ deduces the normal vector $\textbf{n} = [n_x, n_y, n_z]$. Each MLP takes various combinations of inputs, including a 3D position $\textbf{x} \in \mathbf{R}^3$, a unit-norm viewing direction $\textbf{d} \in \mathbf{S}^2$ and an optical centre position $\textbf{o} \in \mathbf{R}^3$. To enrich the inputs with higher-dimensional representations, we employ positional encoding. The encoding $\gamma$ applies to the position $\textbf{x}$ with 9 frequencies to capture spatial variations, while $\zeta$ encodes the optical centre position $\textbf{o}$ using 6 frequencies to encapsulate the optical centre-related nuances. Furthermore, we construct a per-image embedding vector $\ell_i \in \mathbf{R}^k$, and $k$ is the total number of the training images.

\begin{equation}\label{eq:VolumnDensity}
\left\{
\begin{aligned}
& \sigma = F_{\theta\_1} (\gamma(\textbf{x}))
\\
& \textbf{c} = F_{\theta\_2} (\gamma(\textbf{x}), \textbf{d}, \ell_i)
\\
& \textbf{n} = F_{\theta\_3} (\gamma(\textbf{x}), \zeta(\textbf{o}), \ell_i)
\end{aligned}
\right.
\end{equation}

\textbf{Optimization with Depth and Normal Constraint:} During the optimization of the radiance field, we compute the color $\textbf{Ĉ(r)}$ for each pixel through a discretized version of the volume rendering integral. For every pixel, a ray $\textbf{r}(t) = \textbf{o} + t\textbf{d}$ is defined, where $\textbf{o}$ represents the camera's center of projection, $\textbf{d}$ is the viewing direction of the ray, and $t$ is the ray sampling distance.

\begin{equation}\label{eq:8}
\textbf{Ĉ(r)} = \sum_{k=1}^K\omega_k\textbf{c}_k
\end{equation}

Here, $\omega_k$ represents the weight used for accumulating the samples.

We also generate a NeRF depth estimate $\textrm{ẑ}(\textbf{r})$ and a normal estimate $\textbf{\textrm{Ň}(r)}$, and their corresponding standard deviations $\textrm{ŝ}(\textbf{r})$ and $\textrm{ŝ}_{\rm{n}}(\textbf{r})$ to guide the training of the radiance field. These values are derived from the rendering weights $\omega_k$.

\begin{equation}\label{eq:13}
\left\{
\begin{aligned}
& \textrm{ẑ}(\textbf{r}) = \sum_{k=1}^K\omega_kt_k, \quad \textrm{ŝ}(\textbf{r})^2 = \sum_{k=1}^K\omega_k(t_k - \textrm{ẑ}(\textbf{r}))^2
\\
& \textbf{\textrm{Ň}(r)} = \sum_{k=1}^K\omega_k\textbf{n}_k, \quad \textrm{ŝ}_{\rm{n}}(\textbf{r})^2 = \sum_{k=1}^K\omega_k(\textbf{n}_k - \textbf{\textrm{Ň}(r)})^2
\end{aligned}
\right.
\end{equation}

To fine-tune the parameters $\theta\_1$, $\theta\_2$, and $\theta\_3$ of the network, we formulate three distinct loss functions: the color loss $L_{\rm color}$, the depth loss $L_{\rm depth}$, and the normal loss $L_{\rm normal}$. The color loss is only based on a Mean Squared Error (MSE) framework, and the depth loss is solely built on a GNLL term. However, the normal loss is constructed on three different terms, which are a MSE term $l_{\rm n\_MSE}$, a GNLL term $l_{\rm n\_GNLL}$ and an AngMF term $l_{\rm AngMF}$. These distinct terms can improve the normal estimate in different modes, discussed in the following ablation study.

\begin{equation}\label{eq:14}
L_\theta = \sum_{\textbf{r} \in \mathbf{R}}(L_{\rm color}(\textbf{r}) + \lambda L_{\rm depth}(\textbf{r}) + \mu L_{\rm normal}(\textbf{r}))
\end{equation}

\begin{equation}\label{eq:15}
\left\{
\begin{aligned}
& L_{\rm color}(\textbf{r}) = (||\textbf{Ĉ}(\textbf{r}) - \textbf{C}(\textbf{r})||_2)^2
\\
& L_{\rm depth}(\textbf{r}) = 
\mathrm{log}(\textrm{ŝ}(\textbf{r})^2)+\frac{(\textrm{ẑ}(\textbf{r})-Z_{\rm dense\_depth}(\textbf{r}))^2}{\textrm{ŝ}(\textbf{r})^2}
\\
& L_{\rm normal}(\textbf{r}) = l_{\rm n\_MSE} + \eta l_{\rm n\_GNLL} + \tau l_{\rm AngMF}
\end{aligned}
\right.
\end{equation}

The depth loss is applied to rays under two conditions, denoted as $P$ or $Q$, otherwise it will be set to 0. Condition $P$ is met when the absolute difference between the predicted depth $\textrm{ẑ}(\textbf{r})$ and the target depth $Z_{\rm dense\_depth}(\textbf{r})$ is at least as large as the target standard deviation $S_{\rm depth}(\textbf{r})$. Condition $Q$ is satisfied when the predicted standard deviation $\textrm{ŝ}(\textbf{r})$ is equal to or larger than $S_{\rm depth}(\textbf{r})$. The normal loss is also configured in the same way as the depth loss. Implementing the depth loss and the normal loss in this manner enables NeRF to estimate depth values and normal values with an error that falls within one standard deviation of the actual ground truth, enhancing the accuracy of depth and normal predictions.

\subsection{Normal-Assisted Training} 

\textbf{Normal-Improved Distance Sampling:} 
The dense depth prior can provide useful signal to assist distance sampling along a ray, as shown in [14]. Actually, the dense normal prior also contains valuable information to improve distance sampling. Drawing inspiration from GeoNet++[17], we build normal-refine-depth module. This module exploits the normal prior to strengthen the accuracy of the depth prior. First, we suppose that given some 3D points $(x_i, y_i, z_i)$ and their surface normal $(n_{x_i}, n_{y_i}, n_{z_i})$, an unique tangent plane $P_i$ can be determined by the following equation.

\begin{equation}\label{eq:16}
n_{x_i}(x - x_i) + n_{y_i}(y - y_i) + n_{z_i}(z - z_i) = 0
\end{equation}

Based on this plane, we can determine a small neighborhood $M_i$ of pixel $i$. For any pixel $j \in M_i$, we can calculate the depth estimate $z_{ji}$ of pixel $i$ using intrinics $(c_x, c_y, f_x, f_y)$ as follows.

\begin{equation}\label{eq:16}
z_{ji} = \frac{n_{xi}x_j + n_{yi}y_j + n_{zi}z_j}{(u_i - c_x)n_{xi}/f_x + (v_i - c_y)n_{yi}/f_y + n_{zi}}
\end{equation}

After that, we exploit a kernel regression $\textbf{K}$ to aggregate depth value from all pixels in $M_i$ for generating refined depth $\textrm{ẑ}_{ji}$.

\begin{equation}\label{eq:16}
\textrm{ẑ}_{ji} = \frac{\sum_{j}^{M_i} \textbf{K}z_{ji}}{\sum_{j}^{M_i} \textbf{K}}
\end{equation}

Subsequently, the improved depth prior guides the distance sampling along each ray in the same way as [14].

\textbf{Normal Patch Matching:} To improve the fidelity of NeRF's normal outputs, we employ a normal patch match method to evaluate the quality of the rendered normal, expanding upon the ideas presented in [15]. This method is a verification technique that checks the consistency of the rendered normals from multiple views. By using this method, we are able to identify and eliminate normals with excessive noise during the optimization phase of NeRF. This technique is particularly beneficial for CP\_NeRF in assessing the quality of normals in areas lacking texture, where conventional evaluation methods might not be as effective. For areas rich in features, we utilize photometric consistency checks to verify the accuracy of the rendered normals, ensuring their coherence with the observed imagery across different viewpoints. This dual approach allows for a more robust evaluation of normals, enhancing the overall realism and quality of the NeRF-generated scenes.

Before implementing the normal patch match technique, we establish a local 3D plane defined by $\{\textbf{P}|\textbf{P}^T\textbf{n}_q = z_q\textbf{d}^T\textbf{n}_q\}$ for a pixel $q$ in a reference image $I_i$. This plane is situated within the space of the reference camera. In this context, $\textbf{d}$ signifies the view direction, while $z_q$ and $\textbf{n}_q$ denote the distance and the rendered normal from the pixel $q$, respectively. We consider a group of neighboring images to $I_i$, among which is $I_j$. The homography transformation that maps from $I_i$ to $I_j$ can be calculated by (12):

\begin{equation}\label{eq:20}
H = K_j(R_jR_i^{-1}- \frac{(\textbf{t}_i-\textbf{t}_j)\textbf{n}_q^T}{z_q\textbf{d}^T\textbf{n}_q})K_i^{-1}
\end{equation}

Here, $\{K, R, \textbf{t}\}$ represent the intrinsic matrices, rotation matrices, and translation vectors, respectively. This transformation is crucial for applying the normal patch match technique, as it allows us to project the local 3D plane from the reference image $I_i$ onto the neighboring image $I_j$, facilitating the comparison and evaluation of normal consistency across different views.

We utilize the aforementioned transformation to warp a squared patch located on the local 3D plane of pixel $q$ in image $I_i$ to its corresponding position in image $I_j$. To evaluate the visual consistency between the warped patch in $I_i$ and the corresponding patch in $I_j$, we employ the Normalized Cross Correlation (NCC) method. If the NCC value exceeds a predefined threshold, $\varepsilon$, it indicates that the quality of the rendered normal within that patch is insufficient for guiding the NeRF optimization process. In such cases, we assign a weight of zero to the rendered normal in that region, effectively excluding it from the supervision process. This approach is particularly effective in texture-less areas. For regions abundant in textures, relying on the conventional photometric consistency method proves to be more suitable. Finally, the loss function formula (7) is updated as follows:

\begin{equation}\label{eq:14}
L_\theta=\sum_{\textbf{r} \in \mathbf{R}}(L_{\rm color}(\textbf{r})+\lambda L_{\rm depth}(\textbf{r})+\Omega*\mu L_{\rm normal}(\textbf{r}))
\end{equation}

\section{EXPERIMENTS}

Our method is assessed through a baseline comparison and an ablation study on the ScanNet [1] datasets.

\subsection{Experiment Setting}

\textbf{ScanNet:} We utilize COLMAP SfM to derive camera parameters and sparse depth. Based on the sparse depth, we generate corresponding sparse normal. Specifically, SfM is performed on all images to acquire camera parameters. To maintain a clear separation between training and testing data, test images are excluded when generating the point cloud used for rendering the sparse depth maps. On average, the resulting depth maps contain 0.04\% valid pixels. We use three sample scenes, each with 18 to 20 training images and 8 test images. These images exclude video frames with motion blur and ensure surfaces are observed from at least one input view.

\textbf{NeRF Optimization:} Rays are processed in batches of 1024 using the Adam optimizer with a learning rate of 0.0005. To ensure fairness, all methods in the ablation and baseline experiments use 256 MLP evaluations per pixel, irrespective of the sampling approach. The radiance fields are optimized for 500,000 iterations.

\textbf{Evaluation Metrics:} For quantitative assessment, we compute the peak signal-to-noise ratio (PSNR), the Structural Similarity Index Measure (SSIM), and the Learned Perceptual Image Patch Similarity (LPIPS) on the RGB of novel views, along with the root-mean-square error (RMSE) on the expected ray termination depth of NeRF against the sensor depth in meters. Except these, the RMSE on the expected normal of NeRF against the ground truth normal is calculated.

\subsection{Baseline Comparison}

Our method is compared to recent work that also use sparse depth input in NeRF, namely Dense\_Depth\_Priors\_Nerf. The quantitative result indicates that our method surpasses the baseline in all metrics.

"Floaters" are a common issue when applying NeRF with few input views. By using dense depth and normal priors with uncertainty, our method significantly reduces these artifacts compared to the baseline. This leads to much more accurate depth output and greater detail in color. We observed that our method is more robust to outliers in the sparse depth input. This suggests that dense depth and normal priors with uncertainty focus optimization on more certain and accurate views, whereas direct incorporation of sparse depth is more error-prone. Besides greater robustness to outliers, dense depth and normal guide NeRF better at object boundaries not represented in the sparse depth input.

\subsection{Ablation Study}

To validate the effectiveness of the added components, we conduct ablation experiments on the ScanNet scene. The quantitative results show that the full version of our method achieves the highest performance in image quality, depth and normal estimates.

\textbf{Without Normal-Improved Distance Sampling:} Excluding normal-improved distance sampling and using only depth supervision results in inaccurate depth and color due to "floaters" in areas without normal input. Even in areas with dense depth points, the results are less sharp compared to versions that use assisted normal.

\textbf{Without Normal AngMF Loss:} Removing normal AngMF item from the normal loss leads to problems in resolving 
sensitivity against the asymmetric noise in the ground truth surface normal. The quantitative results on ScanNet indicate that considering AngMF item is even more important when using lower quality normal.

\textbf{Without Normal Optical Center Embedder:} Excluding normal optical center embedder results in inaccurate normal and color due to high uncertainty in areas without precise normal input. The lack of normal optical center embedder in the NeRF training leads to lower quality synthesis result.

\textbf{Without Normal Patch Matching:} Excluding the normal patch matching that filters normal error results in an inability to produce precise and consistent normal output across the scene. When rendering novel views, the errors in the dense normal prior become evident, resulting in significant discrepancies between the output and the ground truth.

\begin{table*}
\caption{Comparisons on ScanNet.}
	\label{table:ComparisonsOnScanNet}
	\begin{center}
	\scalebox{1.0}{
	\begin{threeparttable}
	\begin{tabular}{lcccccc}
	\hline
	Method & Img Loss$\downarrow$ & PSNR$\uparrow$  & SSIM$\uparrow$  & LPIPS$\downarrow$  & Depth RMSE$\downarrow$ & Normal RMSE$\downarrow$\\
	\hline
	Baseline   									&0.0086  &20.9657  &0.7136  &0.3202  &0.1838 &N/A\\
	Without Normal-Improved Distance Sampling  	&0.0087  &21.0887  &\textbf{0.7549}  &\textbf{0.3159}  &0.1427 &0.1006\\            
	Without Normal AngMF Loss   				&0.0083  &21.0868  &0.7315  &0.3290  &0.1251 &0.0948\\        
	Without Normal Optical Center Embedder      &0.0078  &21.2123  &0.7445  &0.3254  &0.1255 &0.0881\\
	Without Normal Patch Matching     			&\textbf{0.0076}  &21.3958  &0.7350  &0.3251  &0.1251 &0.0862 \\                
	\textbf{Our Final Method} 					&\textbf{0.0076}   &\textbf{21.4608} &0.7425  &0.3220  &\textbf{0.1237} &\textbf{0.0837}\\ 
	\hline  
	\end{tabular}
	\end{threeparttable}
	}
	\end{center}
\end{table*}

\begin{figure*}
\centering
\includegraphics[width=\linewidth]{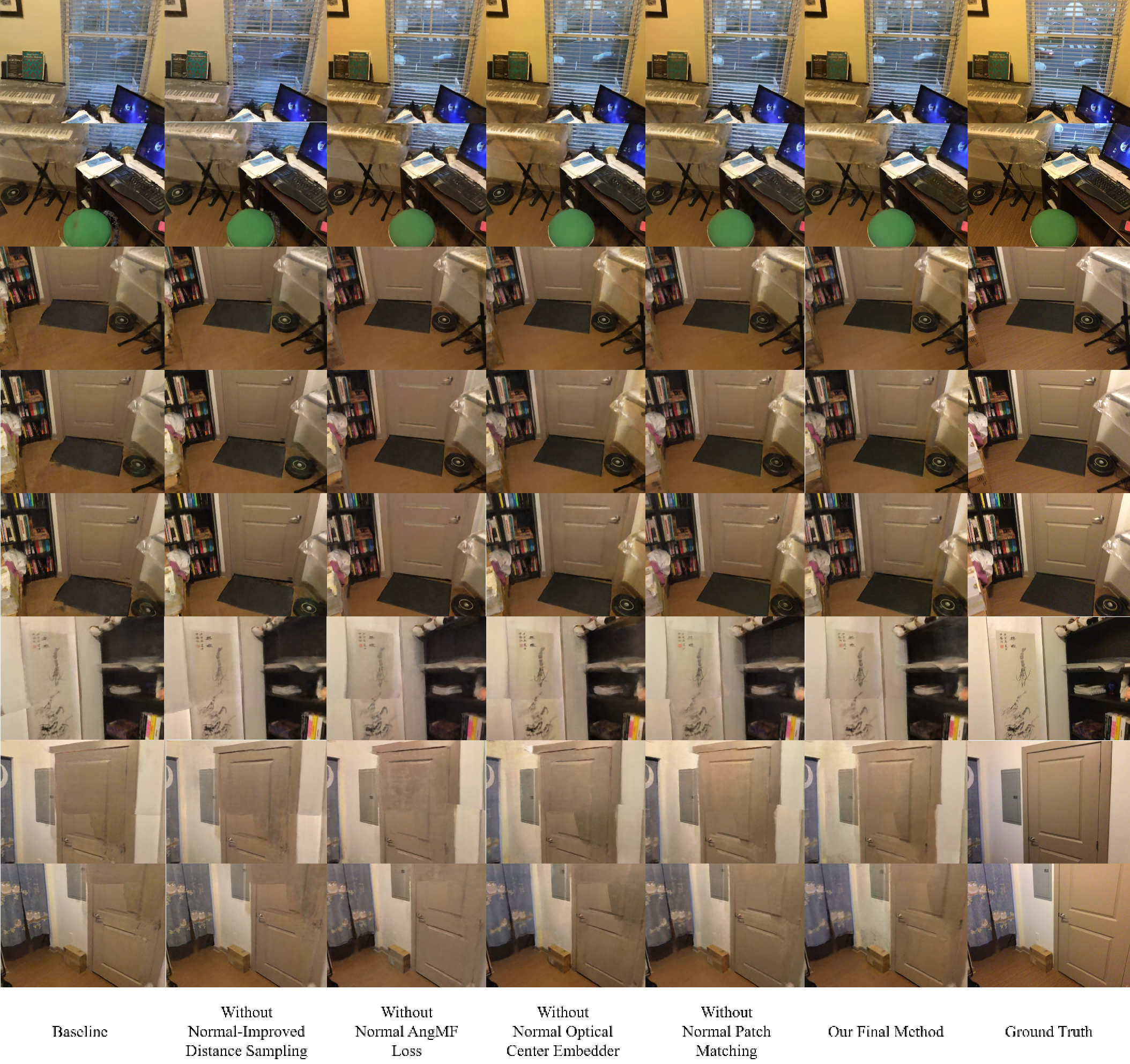}
\caption{The results of the ablation study.}\label{fig:Result1}
\end{figure*}

\subsection{Limitations}

Our method significantly reduces the number of input images required for NeRF-based novel view synthesis while applying it to larger room-sized scenes. However, other NeRF limitations, such as long optimization times and slow rendering, remain. As a result of the drastic reduction in the number of input images, surfaces are typically not observed by more than two other views, limiting view-dependent effects. While our approach optimizes NeRF with as few as 18 images, the depth and normal prior networks require a larger training dataset. Although these priors generalize well and only need to be trained once, it would be beneficial if depth reconstruction could also be learned from a sparse setting.

\section{CONCLUSIONS}

NeRF generate highly realistic images using neural networks but struggle with limited input images, leading to inaccurate renderings. The core issue is the lack of structural detail. To address this, we developed the CP NeRF framework, which enhances NeRF with depth and normal dense completion priors. First, we obtain sparse depth maps via SfM for camera poses and generate sparse normal maps using a normal estimator. These maps train a completion prior with precise standard deviations. During optimization, these priors convert sparse data into dense maps, guiding ray sampling, aiding distance sampling, and constructing a normal loss function for better accuracy. To enhance NeRF's normal outputs, we use an optical center position embedder for more accurate volume-rendered normals and a normal patch matching technique for precise rendered normal maps, ensuring accurate supervision. Our method surpasses leading techniques in rendering detailed indoor scenes, even with few input views.

\addtolength{\textheight}{-12cm}   




\section*{ACKNOWLEDGMENT}

This work was supported by National Key R\&D
Program of China No. U2013202 and the Guangdong Basic and Applied Basic Research Foundation under Grant No. 2022A1515011139.


\end{document}